%% file: bmvc_final.tex
\documentclass{article}
\usepackage{amssymb}
\usepackage{amsmath}
\usepackage{graphicx}

\title{Understanding the Fisher Vector:\\ a multimodal part model}
\date{}
\author{}

\def\eg{\emph{e.g.}}
\def\ie{\emph{i.e.}}

\def\etal{\emph{et al.}}

\usepackage{epstopdf}

\newcommand{\bw}{\mathbf{w}}
\newcommand{\bx}{\mathbf{x}}

\renewcommand{\paragraph}[1]{\par\medskip\noindent{\bf #1}}
\newcommand{\myparagraph}[1]{\par\medskip\noindent{\bf #1.}}

\begin{document}

\begin{center}


{\huge Understanding the Fisher Vector:\\ a multimodal part model}\\[0.8cm]

\noindent
\begin{minipage}[t]{0.5\textwidth}
\begin{flushleft}
David Novotn\'y\textsuperscript{1} \\ 
{\small novotda8@fel.cvut.cz} \\ [0.1cm]
Diane Larlus\textsuperscript{2} \\ 
{\small diane.larlus@xrce.xerox.com} \\  [0.1cm]
Florent Perronnin\textsuperscript{2} \\
{\small florent.perronnin@xerox.com} \\ [0.1cm]
Andrea Vedaldi\textsuperscript{3} \\
{\small http://www.vlfeat.org/$\sim$vedaldi} \\
\end{flushleft}
\end{minipage}%
\begin{minipage}[t]{0.5\textwidth}
\begin{flushleft} 
\textsuperscript{1}Center for Machine Perception \\
\hspace{0.17cm}Czech Technical University \\  [0.2cm]
\textsuperscript{2}XRCE \\
\hspace{0.15cm}Xerox Research Centre Europe \\
\hspace{0.15cm}Meylan, France \\ [0.2cm]
\textsuperscript{3}Department of Engineering Science \\
\hspace{0.155cm}Oxford University \\
\end{flushleft}
\end{minipage}


\end{center}

\begin{abstract}
Fisher Vectors and related orderless visual statistics have demonstrated
excellent performance in object detection, sometimes superior to established
approaches such as the Deformable Part Models. However, it remains unclear
how these models can capture complex appearance variations using visual
codebooks of limited sizes and coarse geometric information. In this work, we propose
to interpret Fisher-Vector-based object detectors as part-based models. Through
the use of several visualizations and experiments, we show that this is a
useful insight to explain the good performance of the model. Furthermore,
we reveal for the first time several interesting properties of the FV, including
its ability to work well using only a small subset of input patches and visual words.
Finally, we discuss the relation of the FV and DPM detectors, pointing out differences
and commonalities between them.
\end{abstract}

\section{Introduction}

Object detection is a key task in computer vision and, as such, 
the community has dedicated to this problem a tremendous amount of effort. 
In the past several years, a predominant line of work in this area has been the use of sliding window classifiers 
computed on top of HOG~\cite{dalal05histograms} or similar features. 
The original HOG classifier of Dalal \& Triggs has since been extended in several ways.
The most representative of such extensions is the {\em Deformable Part Model} (DPM) of Felzenszwalb \etal~\cite{felzenszwalb08a-discriminatively} 
that constitutes today the {\em de-facto} standard solution for a generic object detector. 
The DPM improves on the basic HOG-based detector in two key ways: 
by the introduction of \emph{deformable parts}, allowing templates to deform, 
and by the introduction of \emph{multiple aspects}, 
allowing to capture very different object appearances.

A property of HOG-based models including the DPM, that may explain their popularity nearly as much as their excellent performance, 
is the fact that the learned models are often easily interpretable. 
Already~\cite{dalal05histograms,felzenszwalb08a-discriminatively} used the structure of the HOG features to generate model visualizations 
that convincingly and intuitively capture well-defined object parts. 
More recently, techniques such as HOGgles~\cite{vondrick13hoggles:} 
have been proposed to take these visualizations to the next level, 
revealing many interesting features of these models and of their failure modalities.

HOG-like features are not the only popular image representations in computer vision. 
Before the introduction of HOG, image-based modeling was
flourishing using orderless statistics, starting with the Bag-of-Visual-Words (BoVW)
representation~\cite{csurka04visual,sivic03video}. Similar models are still a
popular choice for image classification and retrieval tasks. In fact, at the
same time as DPM became the most popular model for object detection, orderless
statistics were found to work as well or even better than the
DPMs~\cite{vedaldi09multiple} in international benchmarks such as the PASCAL VOC
challenge~\cite{everingham09voc}. Similar to HOG and DPM, orderless models have ever since been
significantly improved; the best current representative methods include
VLAD~\cite{jegou10aggregating} and, in particular, the \emph{Fisher Vector} (FV)~\cite{perronnin10improving} which, 
compared to BoVW, capture significantly richer statistics of the visual word occurrences. 
Recently, \cite{chen2013emas,cinbis13segmentation} achieved state-of-the-art object detection performance by using FVs.

An open challenge in orderless models is understanding the nature of the visual information that they capture. Part of this challenge is that, differently from HOG, models such as BoVW, VLAD,
and FV are difficult to visualize, so that it is unclear what aspects of an
image or object class they model and how. The reason is that, while HOG
pools local information at well-defined spatial locations in an image, orderless
models scramble this information into a bag, making it difficult to reconstruct
the object being recognized. Furthermore, while in DPM it is easy
to define and visualize a notion of a semantic \emph{part}, the statistical
analogous in the case of BoVW, VLAD, or FV is much less clear. The goal of this
paper is to shed light on these issues by
understanding, interpreting and visualizing these orderless models.
Our focus is the FV representation in
the context of detection, but our conclusions should extend to related models
such as BoVW and VLAD.


Our contributions are threefold. First, we show that the FV detector can be
formulated as a part-based model where what we define as parts have a
similar role to the one of parts in DPMs. Second, based on this new formulation of the
FV detector, we discuss the similarities and the
differences between the FV detector and the DPM, and explain how orderless models
manage to capture varying object appearances using a small visual codebook. In
particular, while DPM
uses unimodal, movable, and well localized parts, the FV detector uses a fixed set of parts,
but is capable of capturing complex multi-modal appearances of each part. Finally
we demonstrate the sparsity of FV detectors, another property shared with DPMs.



The rest of this article is organized as follows. Section~\ref{sec:fv} presents
our FV detection pipeline. Section~\ref{sec:parts} shows evidences
that this model contains parts with a multimodal appearance.
Section~\ref{sec:multimodality} gives insights on the mechanism that allows 
orderless statistics to capture this multimodality into a single model. Section~\ref{sec:sparsity}
evaluates the level of sparsity contained in the model.
Finally, Section~\ref{sec:ccl} summarizes our findings.


\input{fisher_vector}

\section{The Fisher Vector detector as a part-based model}
\label{sec:parts}

A fundamental problem in visual recognition, and especially in object
detection, is accounting for intra-class variability.  
This includes variations in appearance between two views of the same
object instance (\eg difference in view point), but also 
variations in appearance between two different instances of the same
class (\eg the class dog contains instances of Poodles and German Shepards). 
Consequently, images of the same object class 
form a {\em high-dimensional multimodal distribution} which is challenging to model.  

As mentioned earlier, the DPM~\cite{felzenszwalb09object}
improves on the basic Dalal-Triggs~\cite{dalal05histograms} HOG-detector in two key ways: 
by the introduction of \emph{multiple aspects} 
and by the introduction of \emph{deformable parts}.
The use of multiple aspects addresses the multimodal issue:
it allows the model to capture very different object appearances
as caused for example by a large out-of-plane 3D rotation of the object.
The introduction of parts addresses the high-dimensional issue: 
the object is broken-down into smaller ``pieces'' which are easier to model 
because they typically lie in a lower dimensional space. 
The fact that the parts are deformable allows the template to warp geometrically 
and therefore adapt to image-based deformations of the object.
This is important to ensure that the unimodal assumption of each aspect and low-dimensional
assumption of each part are reasonable.
To summarize, the \textbf{DPM is a mixture of aspects} or components, and each aspect is a \textbf{collection of parts}.

The DPM should be contrasted with the FV detector. The latter models the object appearance by extracting $1+R^2$ FVs, one for each spatial subdivision in the pyramid. Each FV captures the appearance of a corresponding spatial bin by pooling local SIFT patches. If we define a part as one of these spatial bins, the model can be seen as a
\textbf{collection of parts} where each part is modeled as a \textbf{mixture of
  Gaussians}. One could argue that individual SIFT patches could also be considered as parts, but Section~\ref{sec:multimodality} shows that these are significantly lower level, closer to part fragments.

In the following, we show that interpreting the FV detector as a part-based model explains how the FV can represent the high-dimensional multimodal appearance of object categories. We start by comparing parts in DPM and FV. Both models have a similar structure: one root part that captures
information at the level of the whole object (the `root filter' in DPM, and the $1\times 1$
spatial bin of the pyramid corresponding to the whole object in FV) 
and several local parts (the `part filters' in DPM, 
and the bins of the $R\times R$ spatial subdivisions in FV).  Yet, their
geometry is different. While parts in DPMs move in order to
match the deforming structure of the underlying object, parts in the FV detector have a rigid predefined layout. A second key difference is that, because of the rigidity of the geometry and the lack of multiple components in the model,
each part in the FV detector is required to capture a highly-variable and multimodal appearance.

The following experiment compares parts in the DPM and the FV detectors, illustrating
the variability of the part appearances captured by the two models. Parts in the FV detector are visualized as follows:
given a class -- motorbike in our example -- and the VOC-07-SMALL test set, the 200 top scoring object detections returned by the FV detector
are selected. The images of these 200 detections as well as the subimages corresponding to the individual parts are extracted and, for each part individually, the corresponding FV descriptors are clustered in six groups using K-means. The average image of each group is shown in Figure~\ref{figure:multimodalityFV}. For comparison, a similar procedure is used to show the parts captured by a DPM model. The 200 top scoring detections are selected and split into
six distinct sets according to the DPM component that was used for each detection. The images belonging to each group are then averaged and shown in Figure~\ref{figure:multimodalityDPM} for the whole object and for each part.


Given Figure~\ref{figure:multimodalityFV} and Figure~\ref{figure:multimodalityDPM}, we note that both models are capable of detecting with high confidence
very dissimilar objects and object parts. However, in the FV detector multimodality is captured at the level of the individual parts, as shown by the highly-variable appearance of the clusters in Figure~\ref{figure:multimodalityFV} (we also tested averaging images without clustering, but this, as expected, blurs any detail). For the DPM, multimodality is captured instead at the level of the components of the mixture, which correspond to meaningful modes in the appearance space.


\newcommand{\CONSTdpmHeight}{1.5cm}
\newcommand{\CONSTfvHeight}{4cm}

\begin{figure} [t!]
\centering
\begin{tabular}{c c}
Whole window (1x1) \vspace{0.1cm}
&
4x4 spatial subdivisions
\\
\includegraphics[height=\CONSTfvHeight]{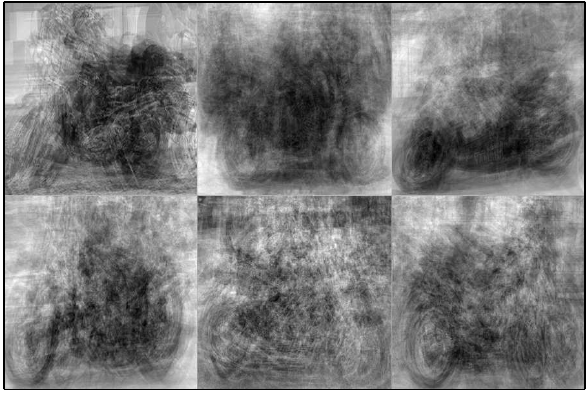}
&
\includegraphics[height=\CONSTfvHeight]{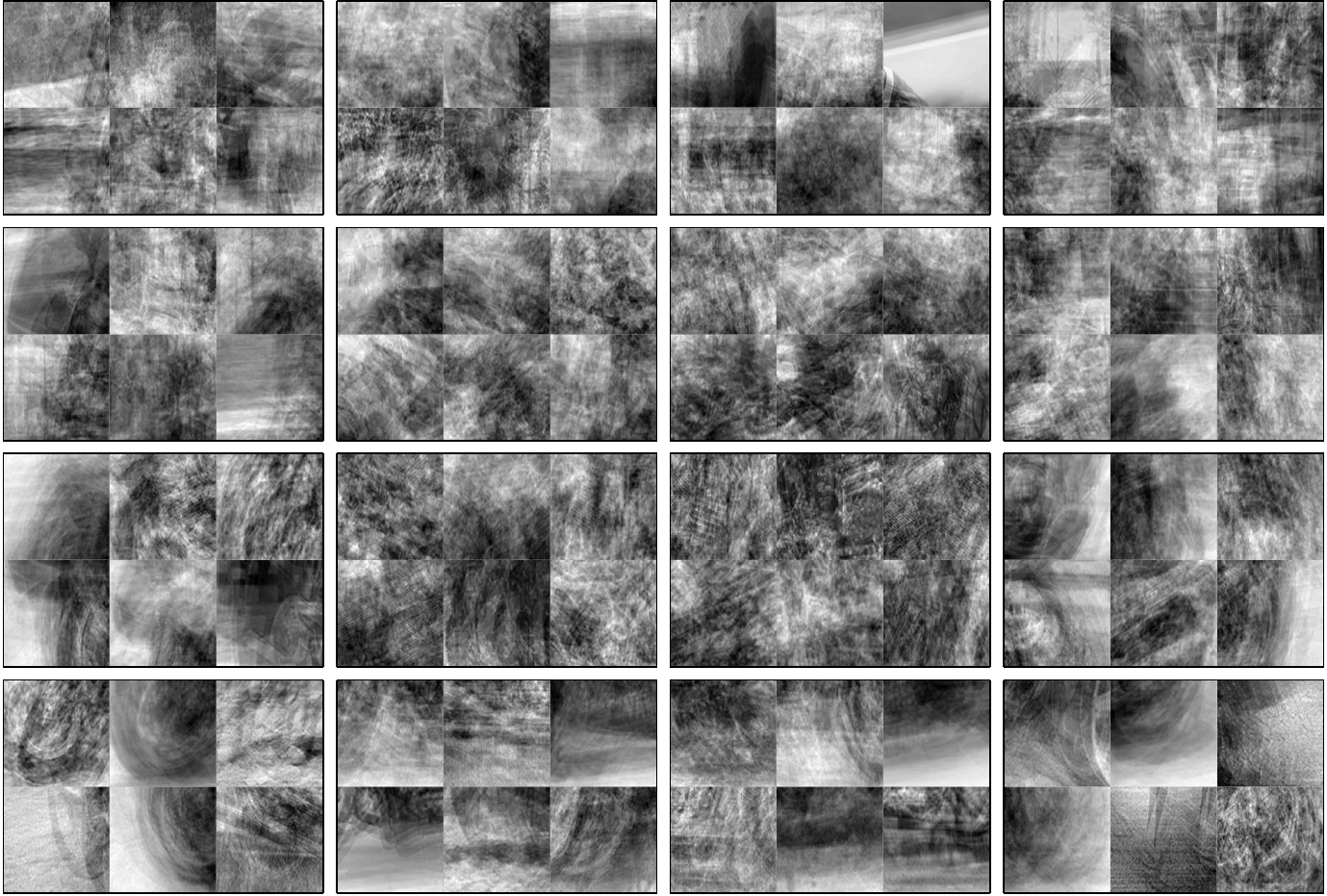}
\end{tabular} \\
\caption{Left: average of the top 200 motorbike detections in VOC-2007-SMALL using the FV detector for the class `motorbike'. Six averages are shown corresponding to clustering the 200 images in six groups, as explained in the text. Right: the same procedure is applied to the individual parts. Note the significant variability of the part appearance.}
\label{figure:multimodalityFV}
\centering
Root filters \\ \vspace{0.1cm}
\begin{tabular}{c c c c c c}
\includegraphics[height=\CONSTdpmHeight]{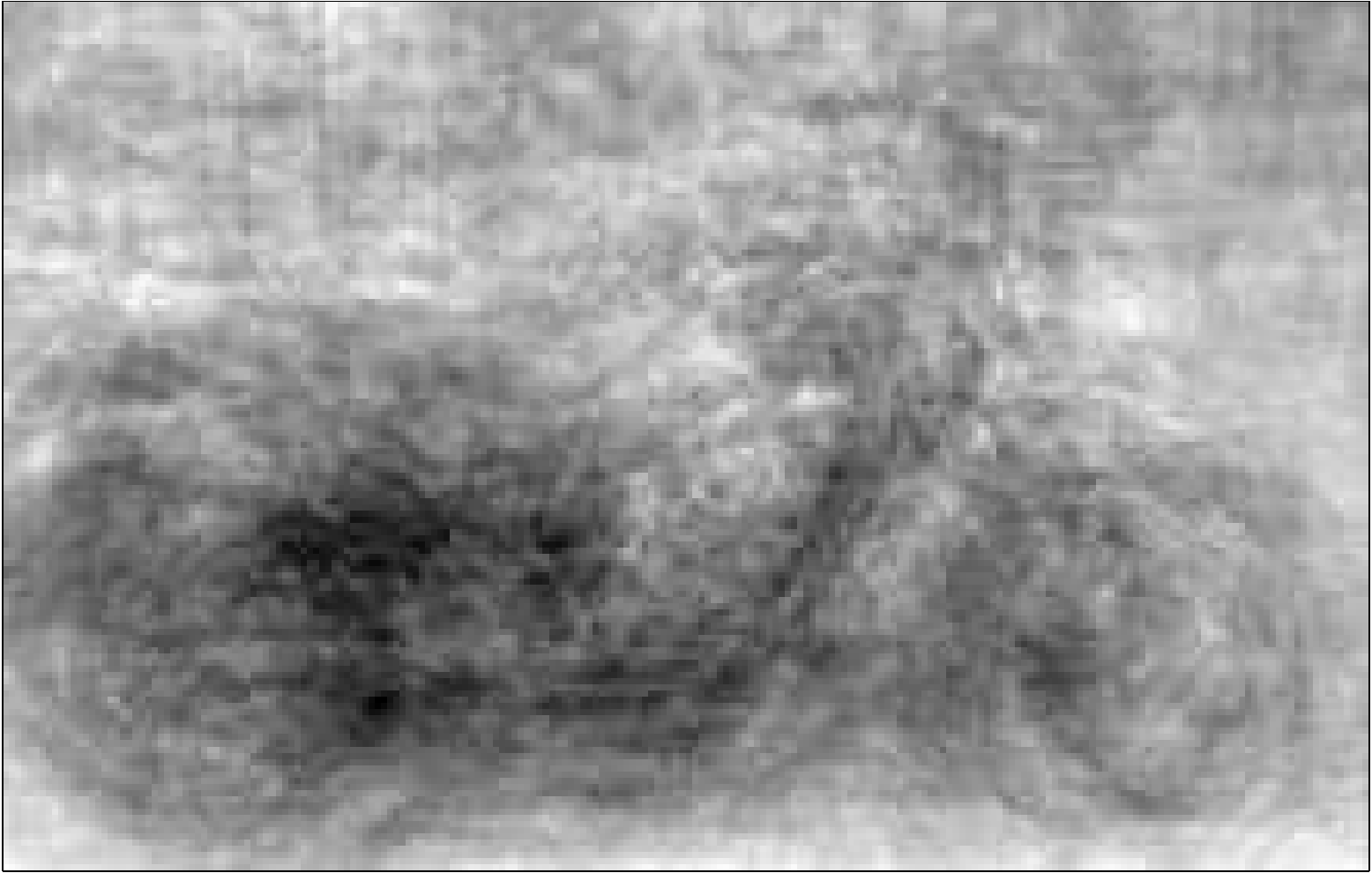} &
\includegraphics[height=\CONSTdpmHeight]{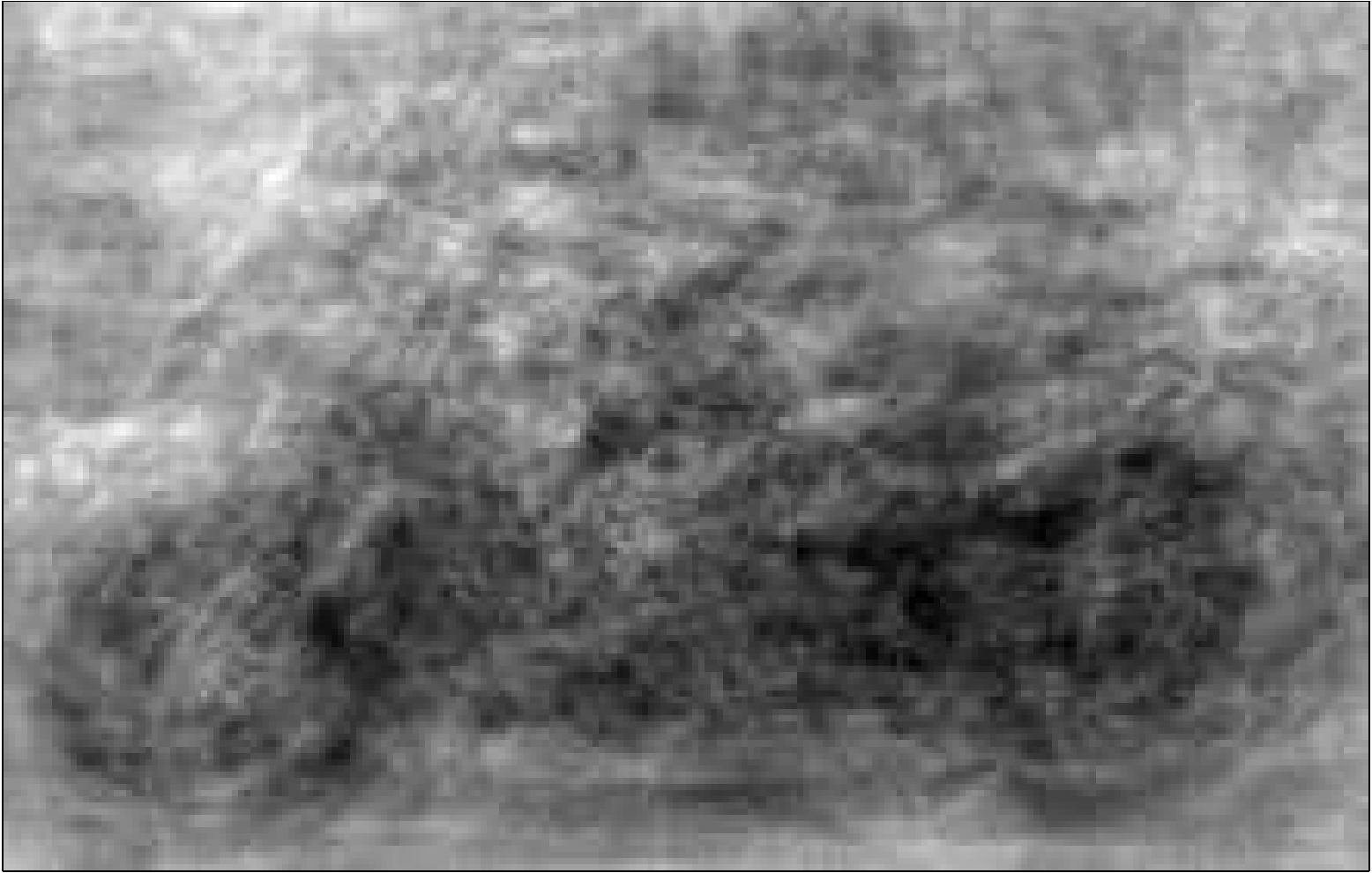} &
\includegraphics[height=\CONSTdpmHeight]{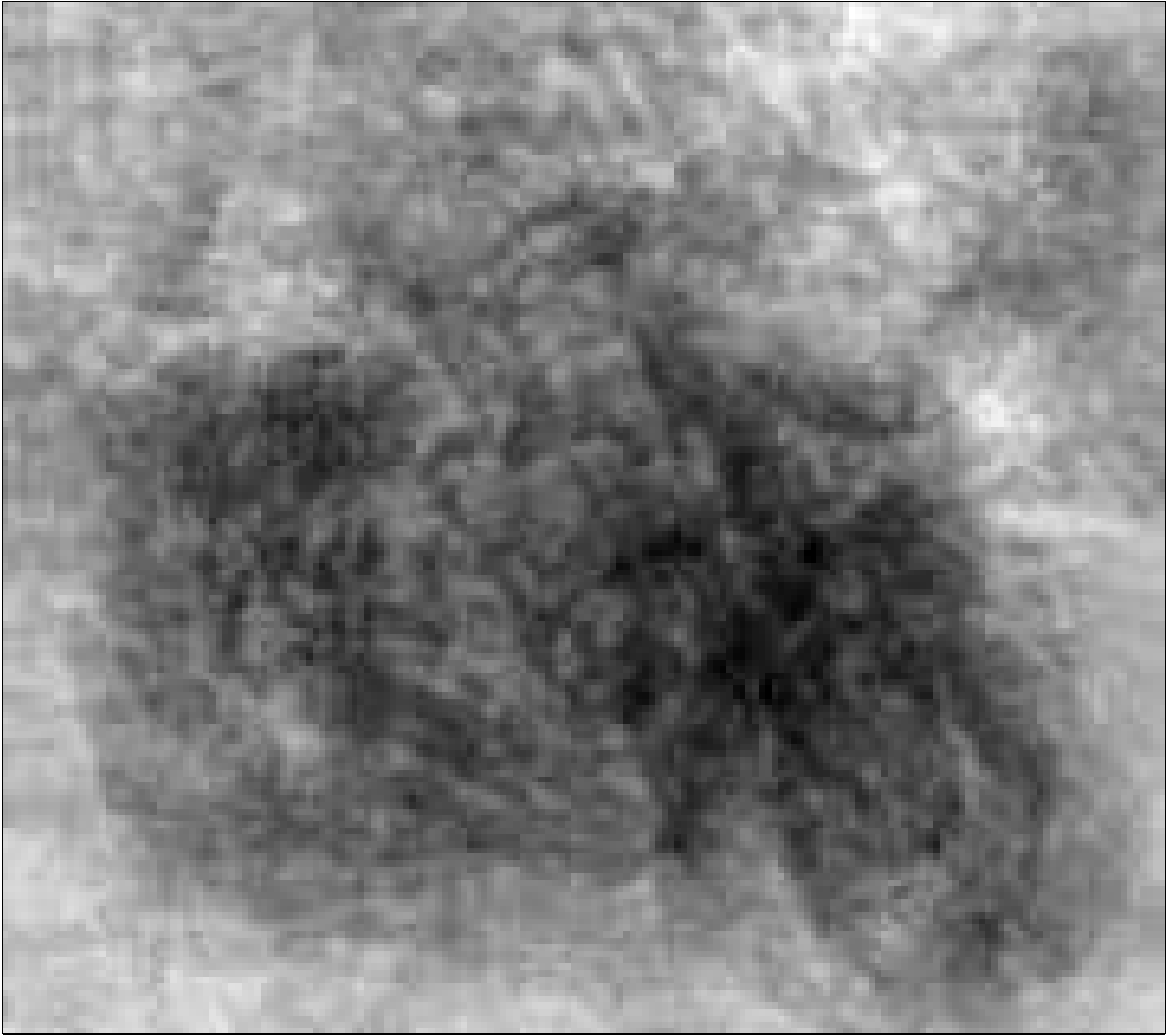} &
\includegraphics[height=\CONSTdpmHeight]{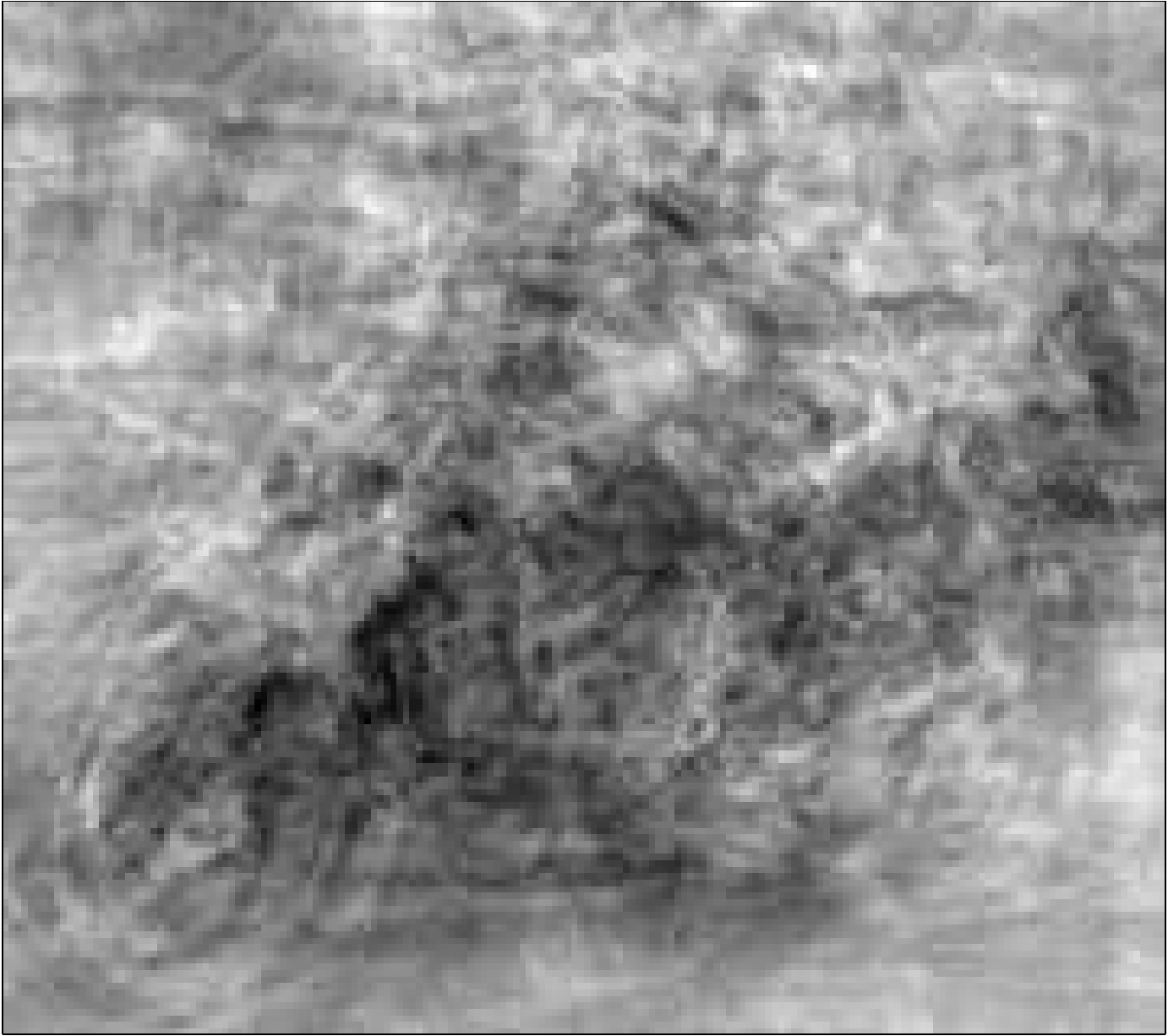} &
\includegraphics[height=\CONSTdpmHeight]{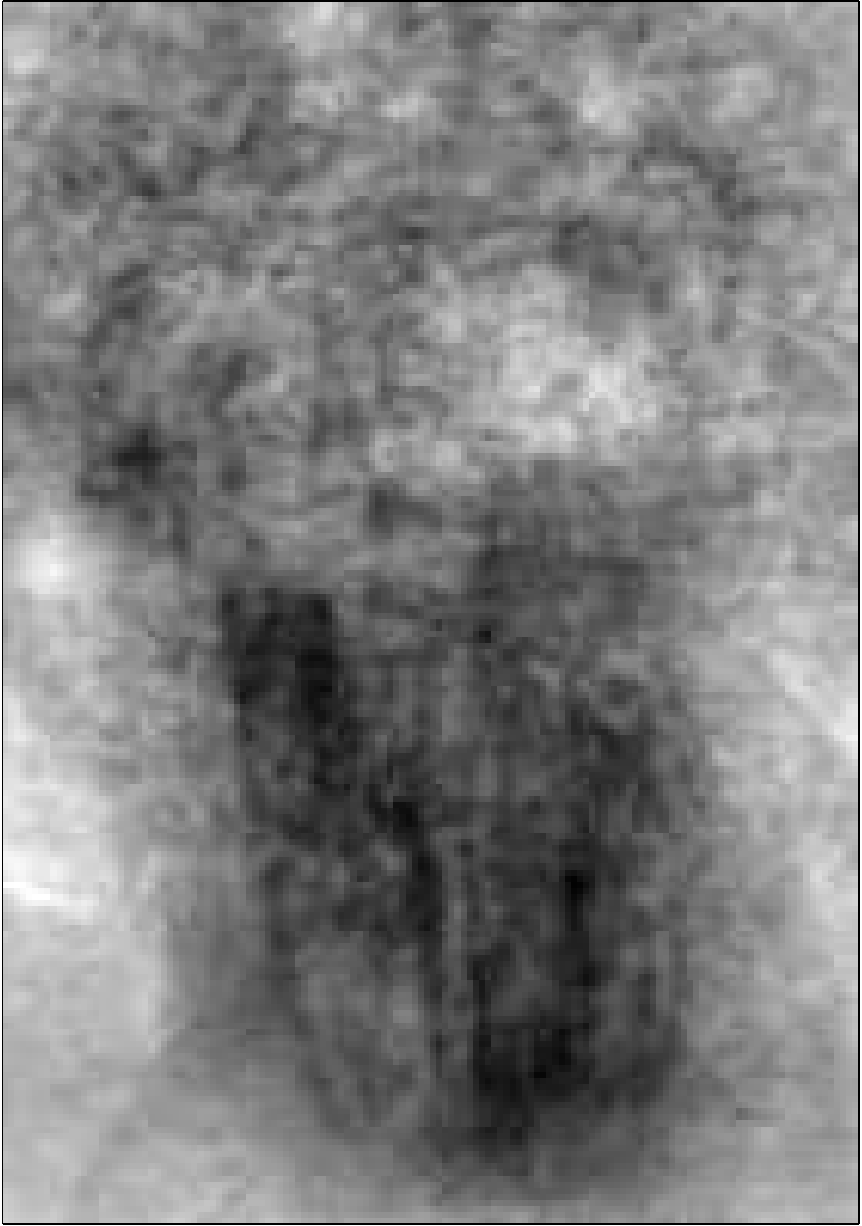} &
\includegraphics[height=\CONSTdpmHeight]{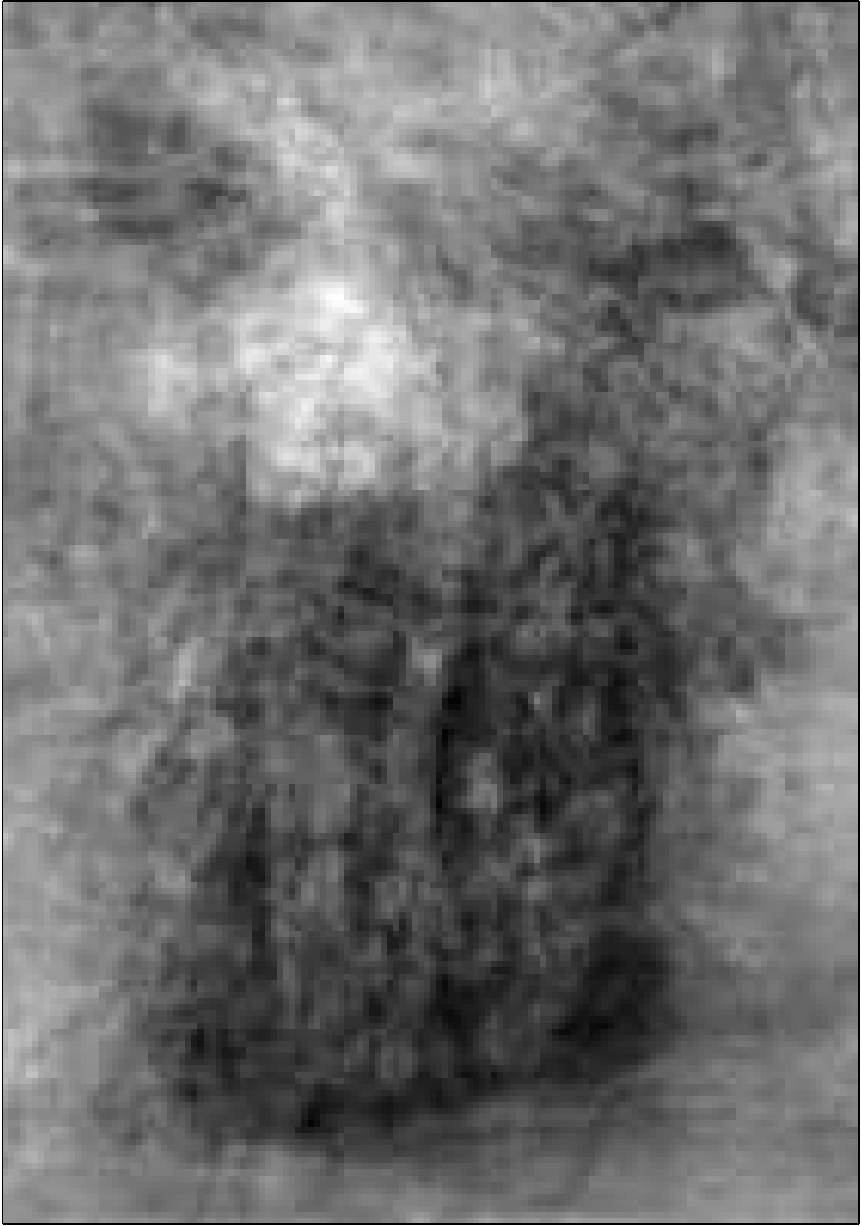}\\
\end{tabular}

Part filters \\ \vspace{0.1cm}
\begin{tabular}{c c c c c c}
\includegraphics[height=\CONSTdpmHeight]{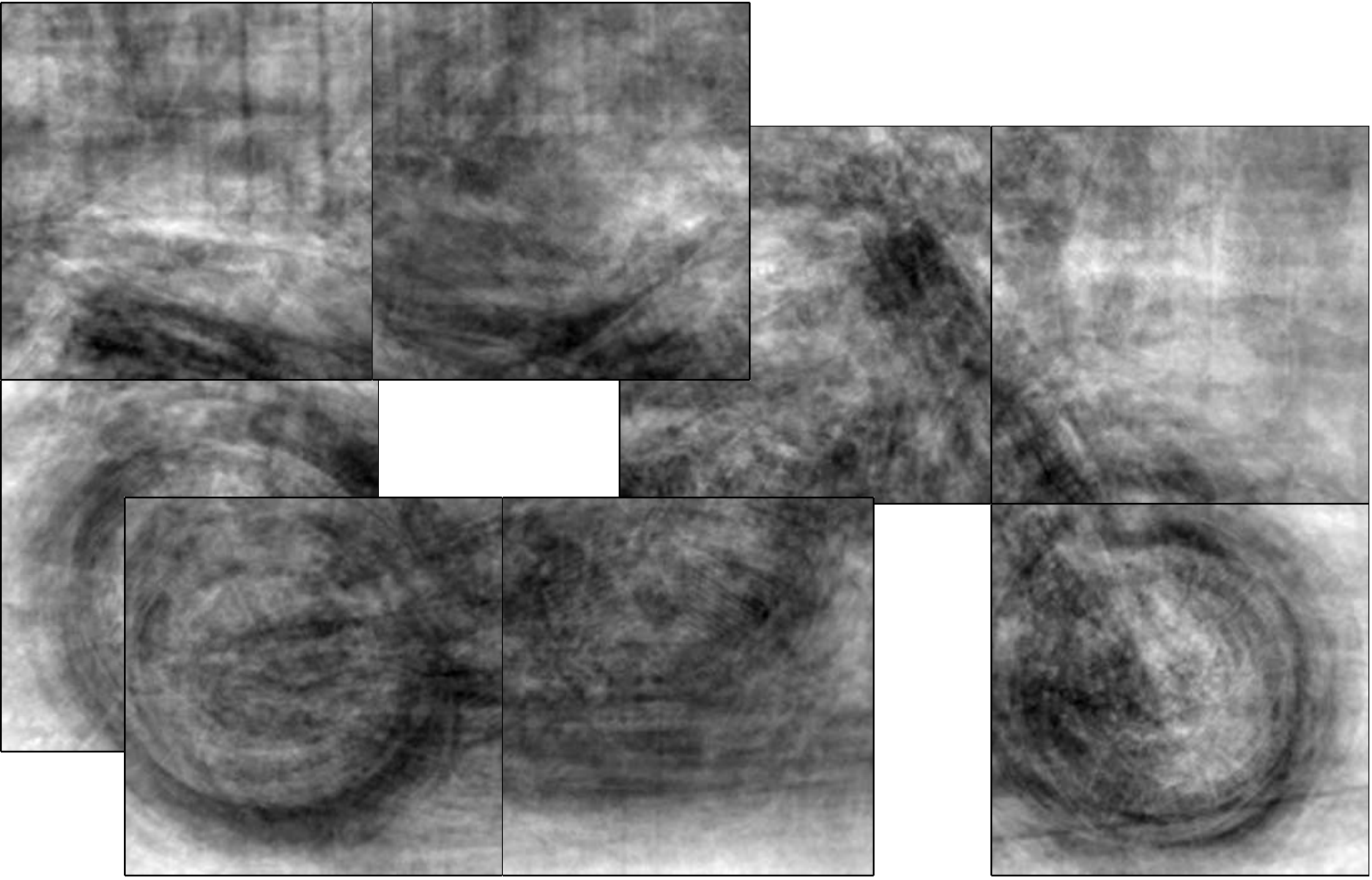} &
\includegraphics[height=\CONSTdpmHeight]{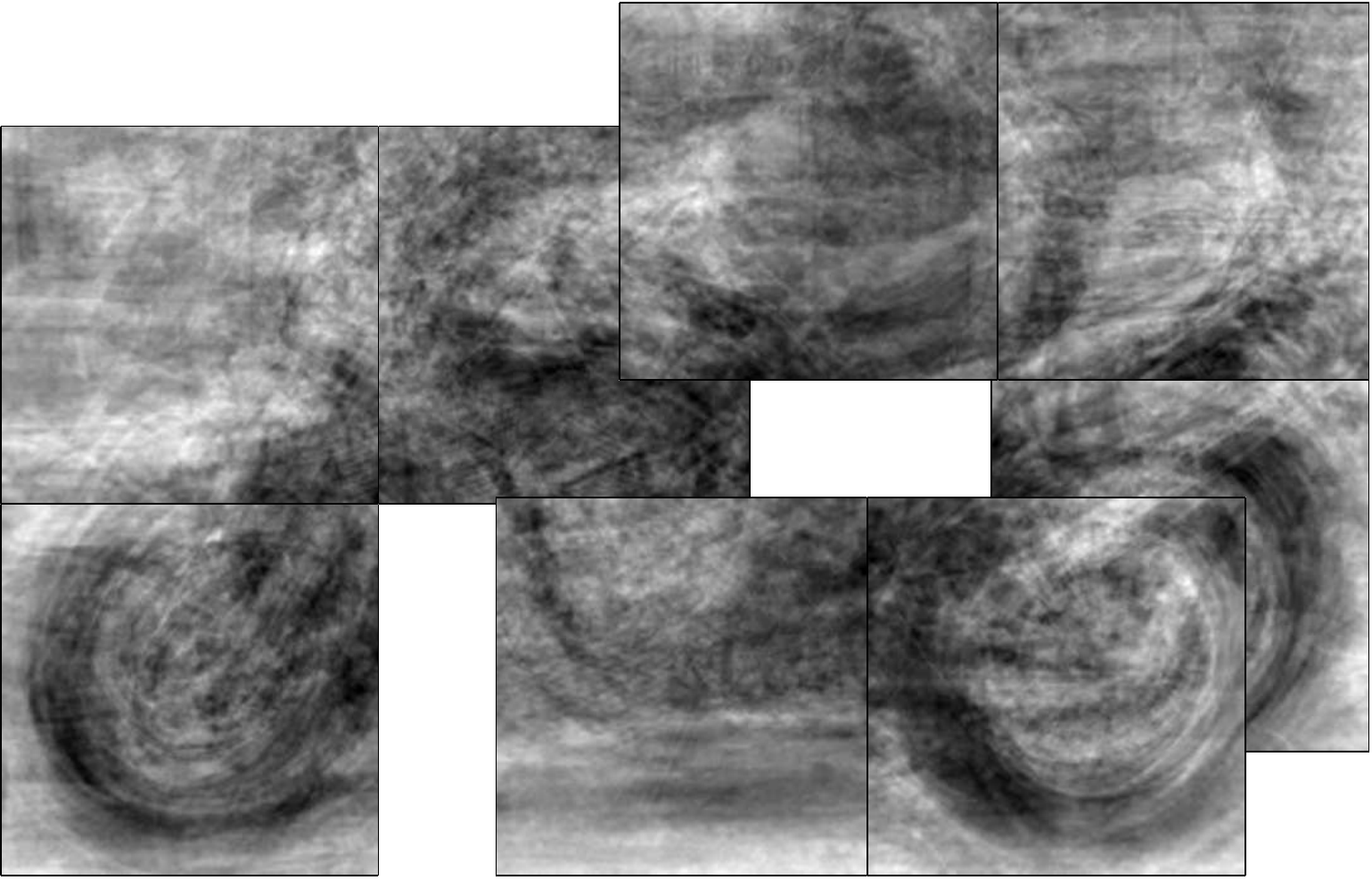} &
\includegraphics[height=\CONSTdpmHeight]{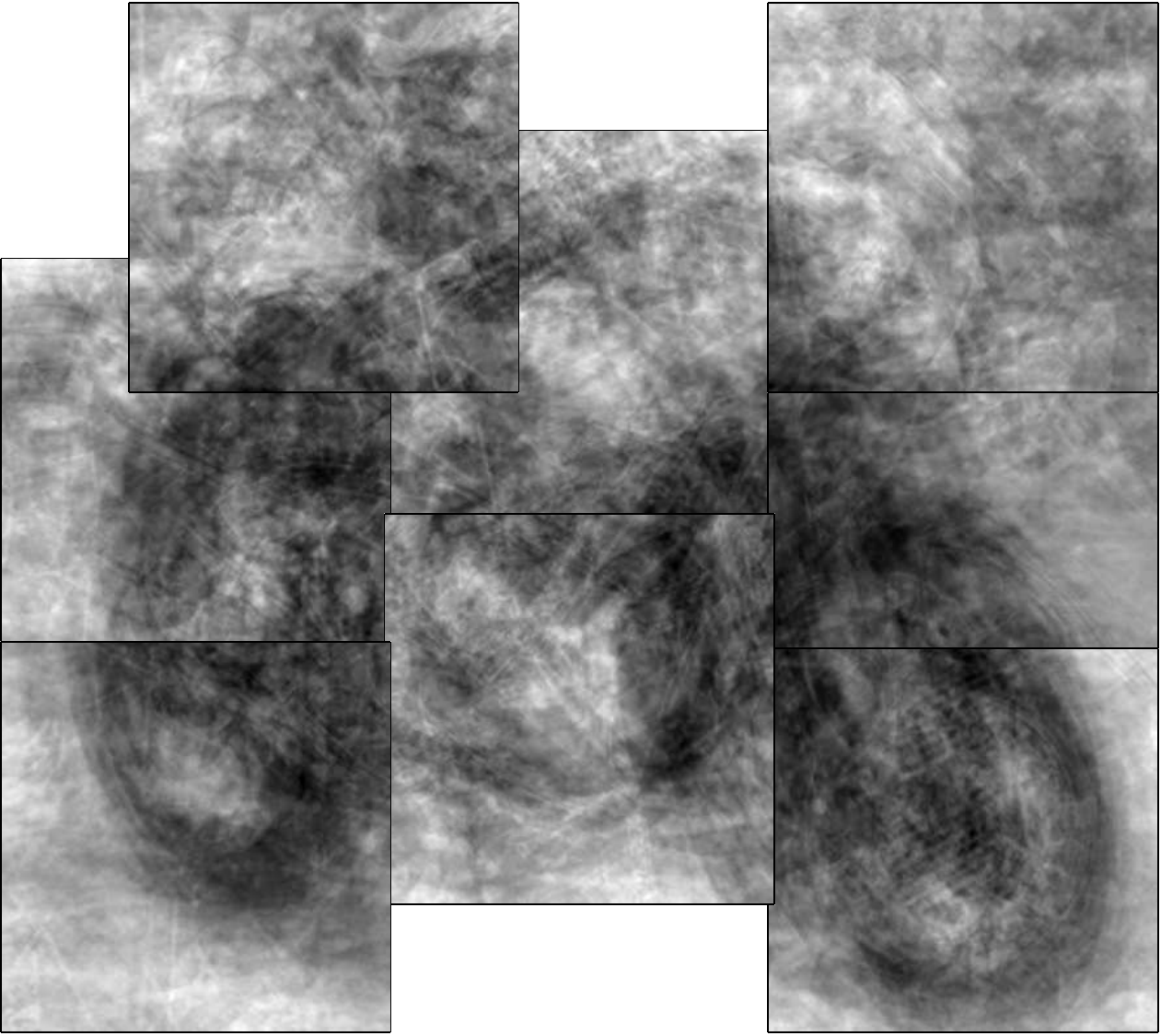} &
\includegraphics[height=\CONSTdpmHeight]{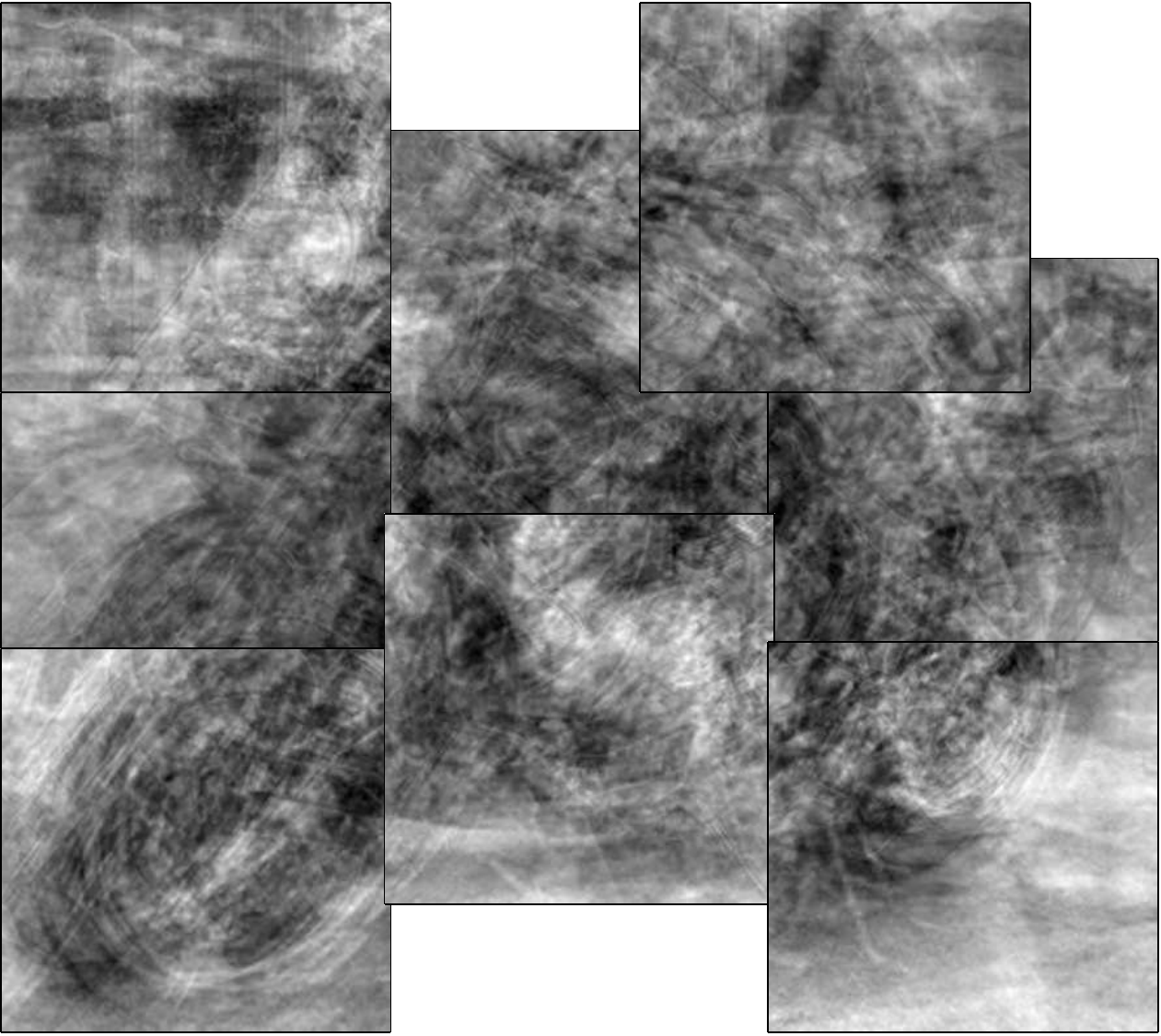} &
\includegraphics[height=\CONSTdpmHeight]{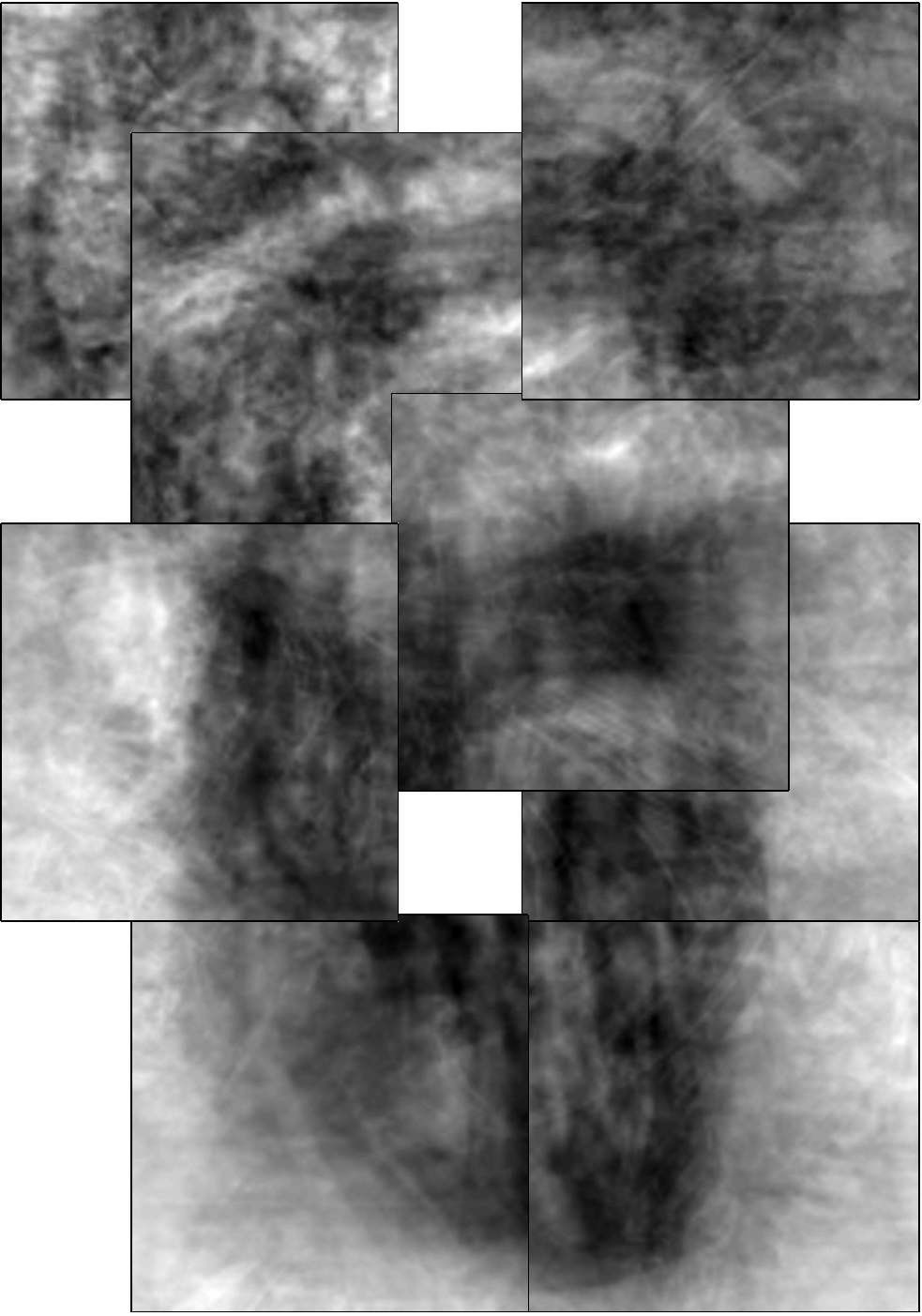} &
\includegraphics[height=\CONSTdpmHeight]{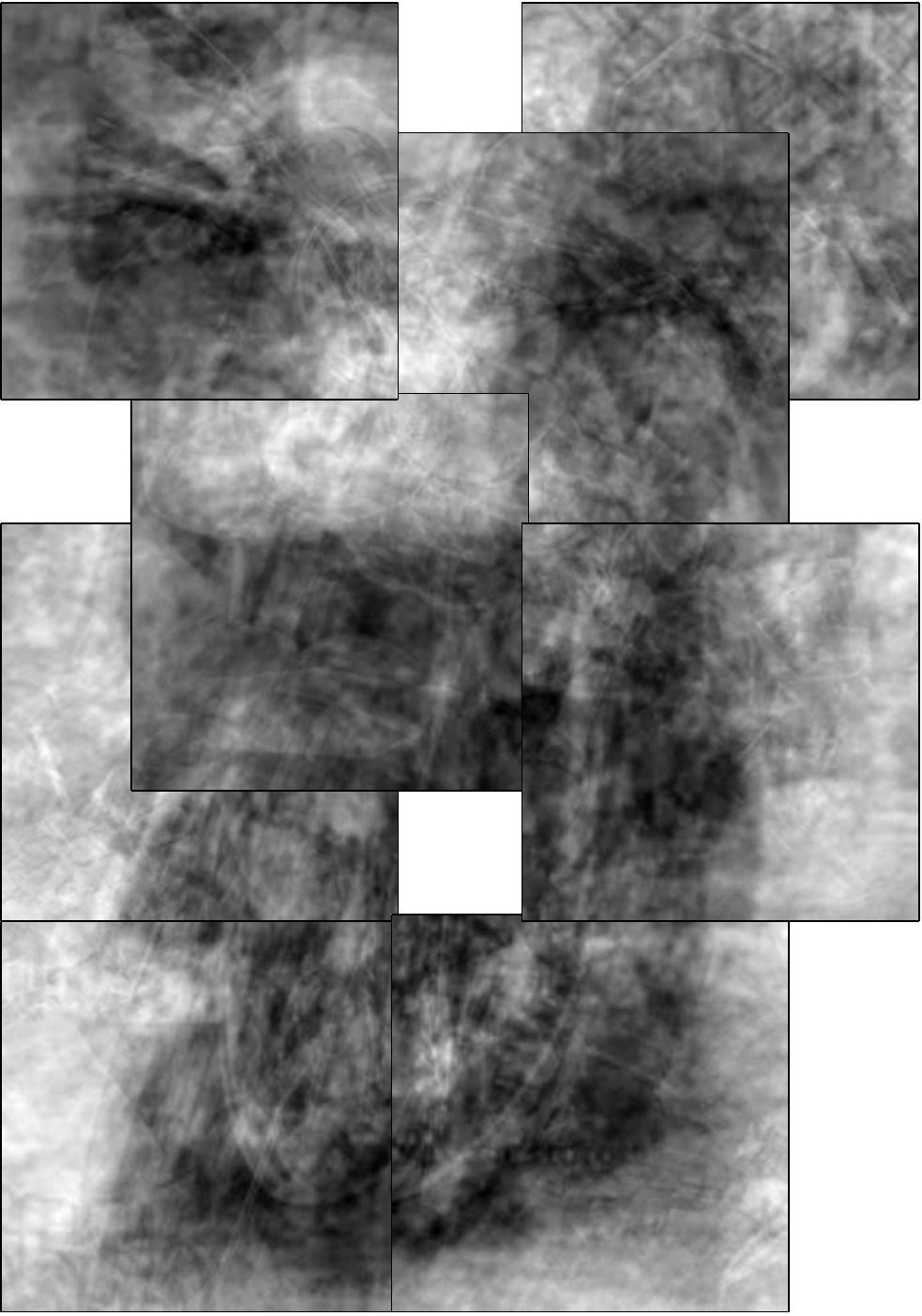}\\
\end{tabular} 

\caption{Average of the top 200 motorbike detections in VOC-2007-SMALL using the DPM model. Six averages are shown, each computed from the subset of the 200 images that activated a particular DPM component.}
\label{figure:multimodalityDPM}
\end{figure}

A direct conclusion of the this observation is that the FV detector 
captures the object variability by means of a
collection of multimodal parts rather than by using a mixture model as the DPM. 
Indeed, if we assume that each part has $M$ modes\footnote{For the simplicity of the analysis,
we assume that we have the same number of modes for each part.} 
and if we have $N$ parts, then the FV detector can model $M^N$ different appearances
of the same object.
In other words, it is possible to model an exponential number of appearances
because the multimodality is factorized per part. 
On the contrary, the DPM can have at most as
many modes as components, because it is built as a mixture model by construction.

Another difference between the DPM and FV detectors lies in training.
The DPM learns a different part-based model for each component,
where components are roughly divided according to viewpoints. In practice, components are initialized based on aspect-ratio or another heuristic, and eventually assembled in a mixture model using latent variables. In contrast, the FV detector learns in one shot a single linear classifier capturing the whole space of object appearances; nevertheless, the factorization property of the FV allows this procedure to capture efficiently an exponential number of different appearances.


\section{Modeling multimodal parts with the Fisher Vector}
\label{sec:multimodality}

In the previous section we have interpreted the FV detector as a
part-based model, and we have given evidence that these parts are highly
multimodal, a necessary property for dealing with the highly variable appearance
of object categories. In this section we investigate the mechanism that allows FV
to capture such rich multimodal appearances. In BoVW models, multimodality is
easily explained as the representation quantizes the feature spaces in thousands
of different visual words. However, the quantization granularity is much smaller
for FV, typically in the order of a few dozens Gaussian components. 
This section clarifies why such a small number of visual words is sufficient to represent rich appearance variations.

The first answer to this question lies in the statistics encoded by the
FV. While both BoVW and FV quantize local patch descriptors using a visual codebook, BoVW captures only
$0^\text{th}$-order statistics (counts) of the features, while FV captures first order and
second order statistics as well (see Eq.~(\ref{eq:fv})). In particular, the point-wise FV $\phi(x_i)$ is a mixture of quadratic functions of the descriptor $x_i$, where different functions are activated based on the descriptor quantization. Hence, a linear classifier $\langle \bw, \Phi(x_i) \rangle$ learnt on the FV representation can be seen as a mixture of quadratic experts in the space of local patch descriptors.

Our next experiment demonstrates the expressive power of the FV representation by showing that a linear classifier learned on top of FV can induce complex decision boundaries in descriptor space.
To this end, we consider a patch $x_i$ and encode it using the point-wise FV $\Phi(x_i)$. We then associate
this patch $x_i$ to a score $s(x_i|c,b) = \langle \bw_{cb}, \frac{\Phi(x_i)}{\| \Phi(x_i) \|_2}  \rangle$ where
$\bw_{cb}$ denotes the weight vector learned by the FV detector for class $c$
and spatial bin $b$.
For the purpose of this visualization, we score a single patch at a time
although the weights $\bw_{cb}$ are trained on the full FV model that pools information from all the patches covering an object. While in this manner we cannot visualize the aggregated effect of all the patches, Sect.~\ref{sec:sparsity} shows that only a small number of those is actually important for classification making this a good proxy.

\begin{figure}[t]
\hfill
\includegraphics[width=3.5cm]{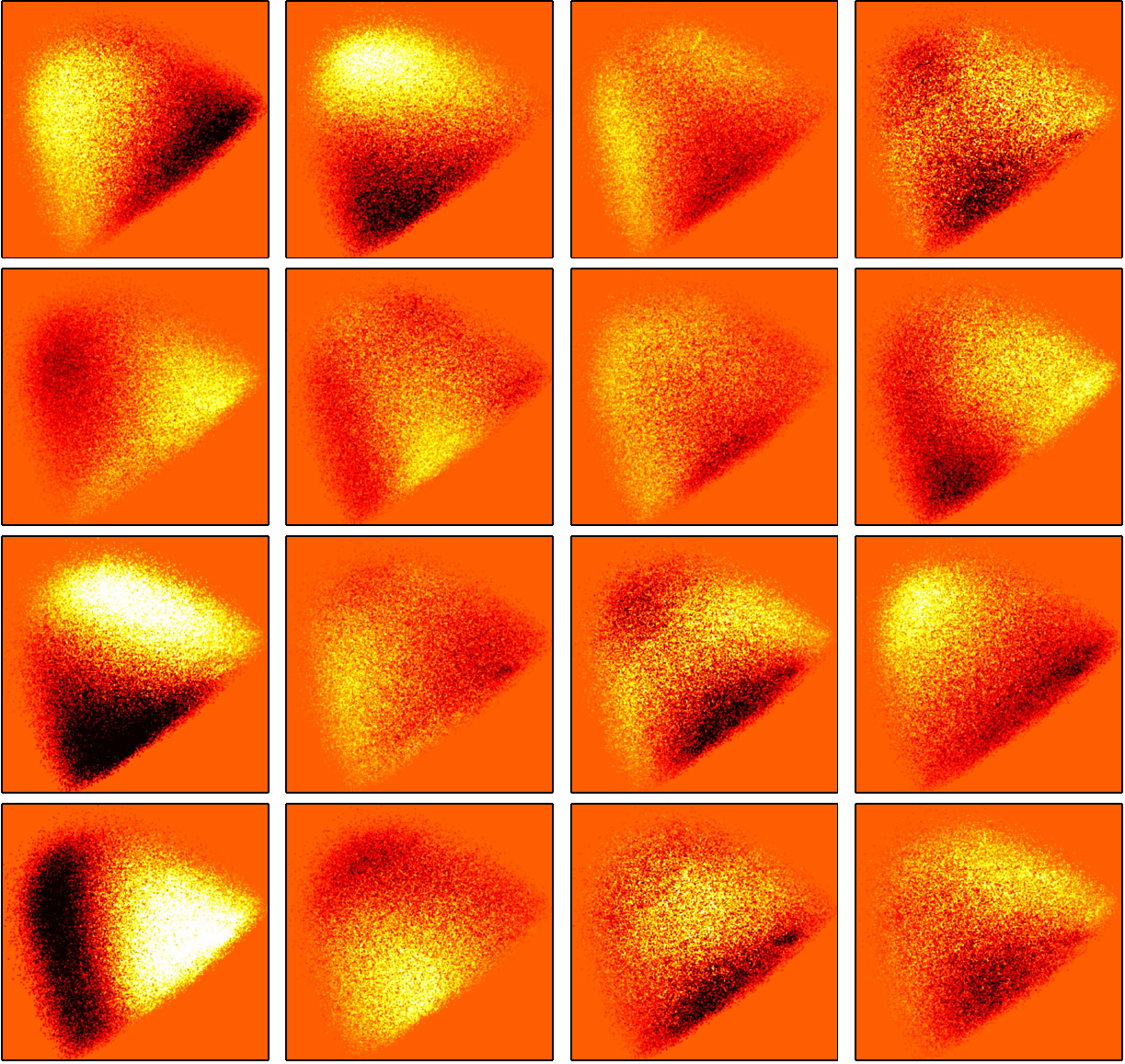}\ (a)
\hfill
\includegraphics[width=3.5cm]{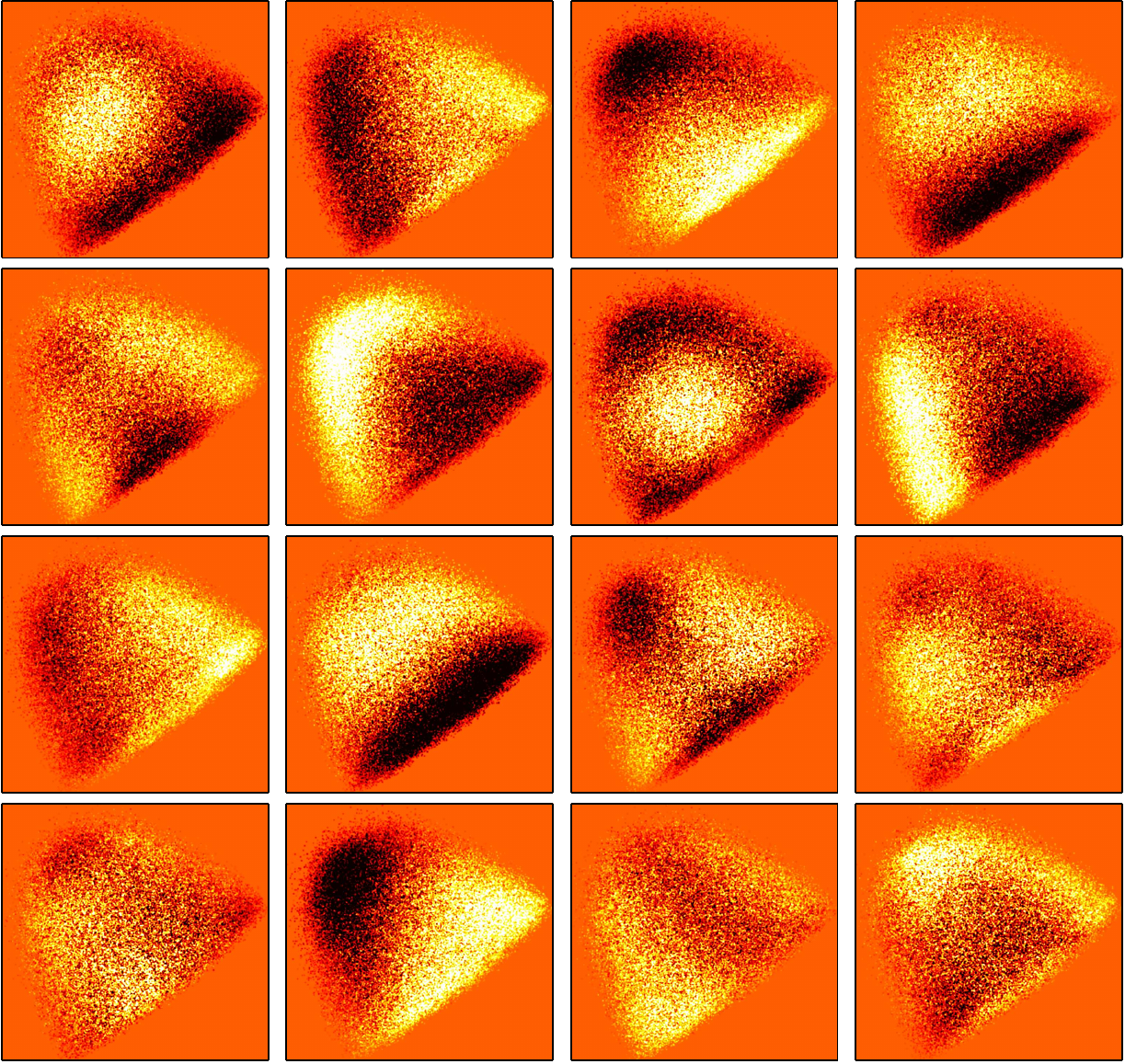}\ (b)
\hfill
\includegraphics[width=1.4cm]{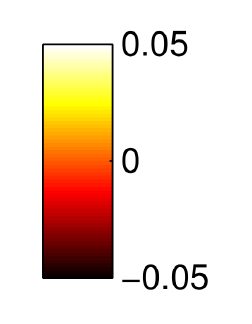}
\hfill
\caption{The FV can be seen as a mixture of quadratic experts in the space of local descriptors. The figure illustrates the variety of scoring functions that a single expert, corresponding to a single Gaussian in the model, can express. Each image shows the scoring functions obtained for each of the $4 \times 4$ spatial bins of (a) buses and (b) motorbikes.}
\vspace{-2mm}
\label{figure:boundary}
\end{figure}

Our goal is to illustrate the richness of the scoring functions  $s(x_i|c,b)$ that
the FV representation associates to local descriptors even when a single
Gaussian mode is considered. To this end, we restrict the function domain to the
patches $x_i \in X_k$ that are assigned to a certain Gaussian $k$. We then plot the function $s(x|c,b): x \in X_k$ for
different categories $c$ and spatial bins $b$. Since the input of this function
are 128 dimensional SIFT descriptors ($X_k \subset \mathbb{R}^{128}$), we first
parametrize the descriptor space with a 2D space $\hat{x} = M_k x,$ $\hat x \in
\mathbb{R}^2$ where $M_k$ is a PCA projection learnt from a set of descriptors
sampled from $X_k$. Figure~\ref{figure:boundary} shows $s(M_k x|c,b)$ for a Gaussian $k$, two classes $c$ (bus and motorbike), and $4\times 4=16$ spatial bins. These results are representative of other classes and Gaussian components.

There are two points to take from the visualizations in Figure~\ref{figure:boundary}. First, scores are well clustered in a small number of modes, due to the smooth form of the scoring function and of the encoding function $\Phi(\bx_i)$. Second, the modes are nevertheless very varied, both for different classes and for different spatial bins, showing that the same Gaussian cluster has significantly different meaning depending on the class as well as on the spatial location.
The ability of ``reusing'' the same visual words to express varied decision functions explains how the FV is able to capture complex multi-modal object appearances while using visual vocabularies significantly smaller than BoVW.

\begin{figure} [t]
\centering
\begin{tabular}{c c c c}
\includegraphics[width=2.5cm]{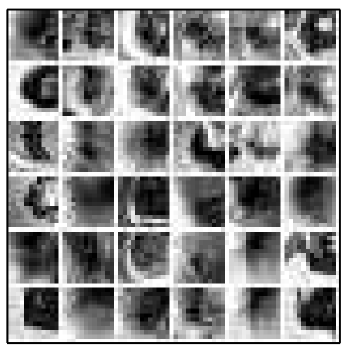} &
\includegraphics[width=2.5cm]{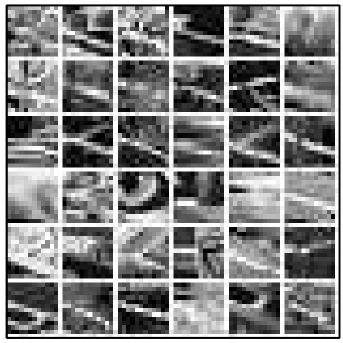} &
\includegraphics[width=2.5cm]{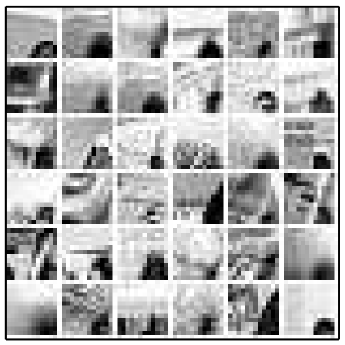} &
\includegraphics[width=2.5cm]{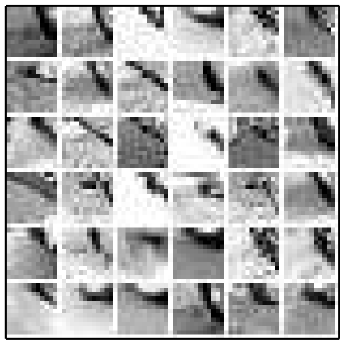} \\
\includegraphics[width=0.5cm]{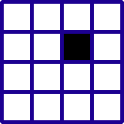} & 
\includegraphics[width=0.5cm]{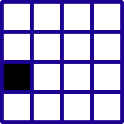}  &
\includegraphics[width=0.5cm]{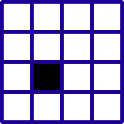}  &
\includegraphics[width=0.5cm]{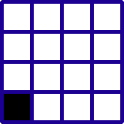} \\
(a) & (b) & (c) & (d)
\end{tabular}
\caption{Highest scoring patches for some Gaussians and some
  spatial bins (bottom row) of the $4\times 4$ spatial layout, for the cat
  (a), sheep (b), bus (c) and motorbike (d) classes.}
\label{figure:boundary2}
\vspace{-2mm}
\end{figure}

While Figure~\ref{figure:boundary} shows that the FV is capable of assigning complex scoring functions to local patches, it does not clarify whether the learned scores are ``semantic'', in the sense of identifying interpretable image fragments. Our next goal is to investigate this question. To do so we consider the set $X_k$ of all image patches extracted from the
VOC-07-SMALL dataset and assigned to the $k^\text{th}$ Gaussian component of the model. Within this set, we select the 36 patches $x_i$ that maximize the score $s(x_i|c,b)$ for a given category $c$ and spatial bin $b$. We visualize these patches in
Figure~\ref{figure:boundary2} for different objects and spatial bins.
Despite the fact that all the patches in Figure~\ref{figure:boundary2} happen to belong to
true positive detection windows, it is very hard to recognize which object parts they belong to. Hence, we conclude that the local patches pooled by the IFV are akin to \emph{part fragments} rather than semantic parts; on the other hand, Figure~\ref{figure:multimodalityFV} suggests that, by aggregating many of these fragments together, the spatial bins in the FV can recognize meaningful parts.


We conclude that the FV captures multimodal part appearances by (1) decomposing parts into distributions of lower-level image fragments and (2) by learning complex scoring functions for these fragments despite the use of a small visual vocabulary.


\section{Sparsity properties of the Fisher Vector detector}
\label{sec:sparsity}

A desirable property of the DPM -- at least on a visualization standpoint, but
to some extent also on a computational point of view -- is that it is sparse, in
the sense that it captures the appearance of objects using a small set of part
templates that fire in correspondence of selected image regions that are aligned
to the parts (a consequence of the use of max-pooling). On a first
glance, the FV detector does not seem to exhibit any sparsity property: patches
are quantized using a GMM with a number of components that, while much smaller
than visual words in BoVW, is still larger than the number of parts in a DPM;
furthermore, thousands of image patches are encoded and averaged due to the use
of sum-pooling instead of max-pooling. We revisit this assumption and study the
sparsity properties of the FV detector at two different levels, the patch level
and the Gaussian level.



\myparagraph{Sparsity of the patches}
Sect.~\ref{sec:multimodality} and Figure~\ref{figure:boundary} looked at the scores associated by point-wise Fisher Vectors to individual top-ranked image patches. However, the FV detector pools information from hundreds of patches in the detection window and it is unclear whether the final decision depends on these relatively rare highly-scoring patches or, instead, the majority of other patches that receive intermediate scores. In other words, we do not know whether information is concentrated as it happens in the DPM case or instead distributed uniformly in the detection window. The next experiment answers this question.

We start from the hypothesis that top-ranked patches (\ie those at the centers of the positive and negative modes in Figure \ref{figure:boundary}) 
contain most of the information necessary for detection. In order to validate this hypothesis, we drop from each detection window a certain portion of patches, starting from the ones with smallest absolute individual score, while
keeping track of the achieved APs. Intuitively, if these low-scoring patches have a small effect on the final classifier score, the detection APs should be stable. However, once the important patches start to be
removed APs should decrease rapidly. 

\begin{figure}[t]
\centering
\label{figure:gaussianSparsity}
\centering
\begin{tabular}{c c}
\includegraphics[width=0.49\linewidth]{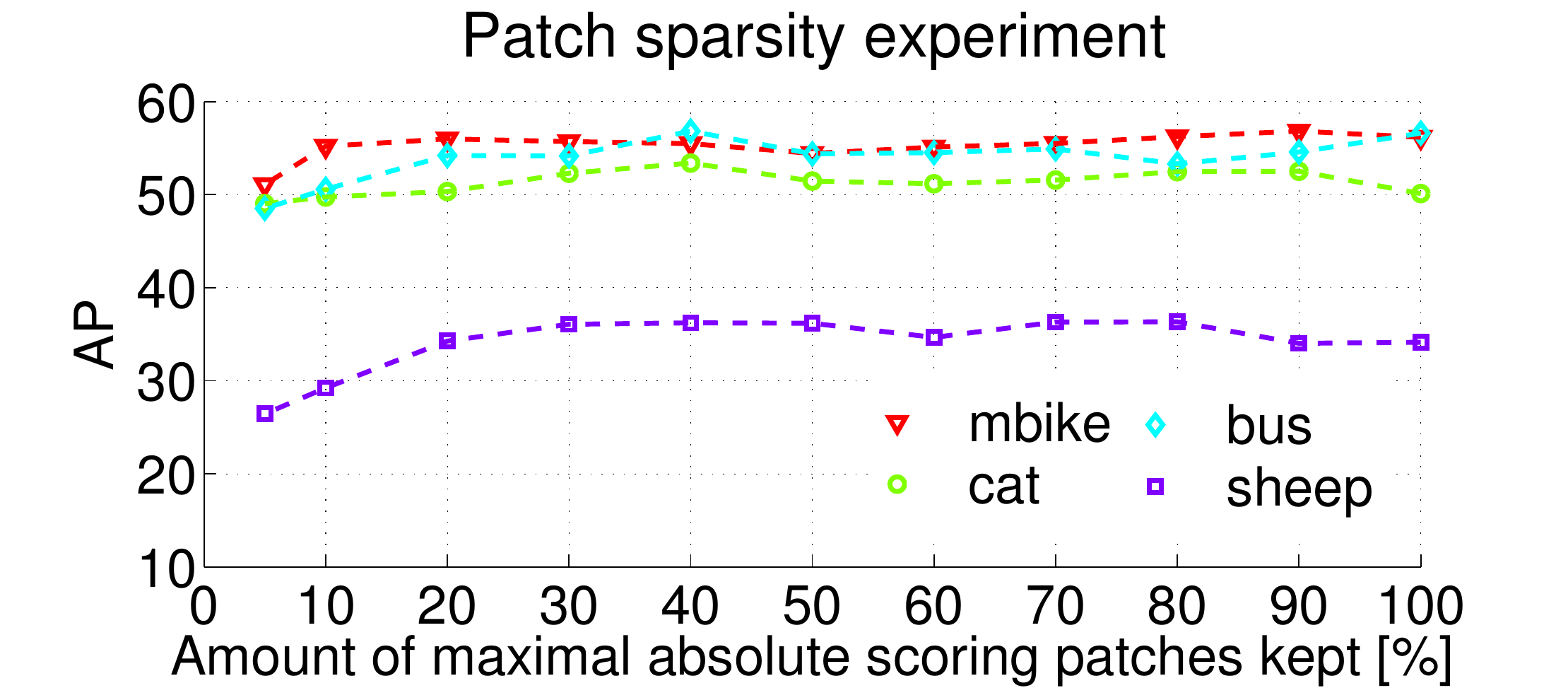} &
\includegraphics[width=0.49\linewidth]{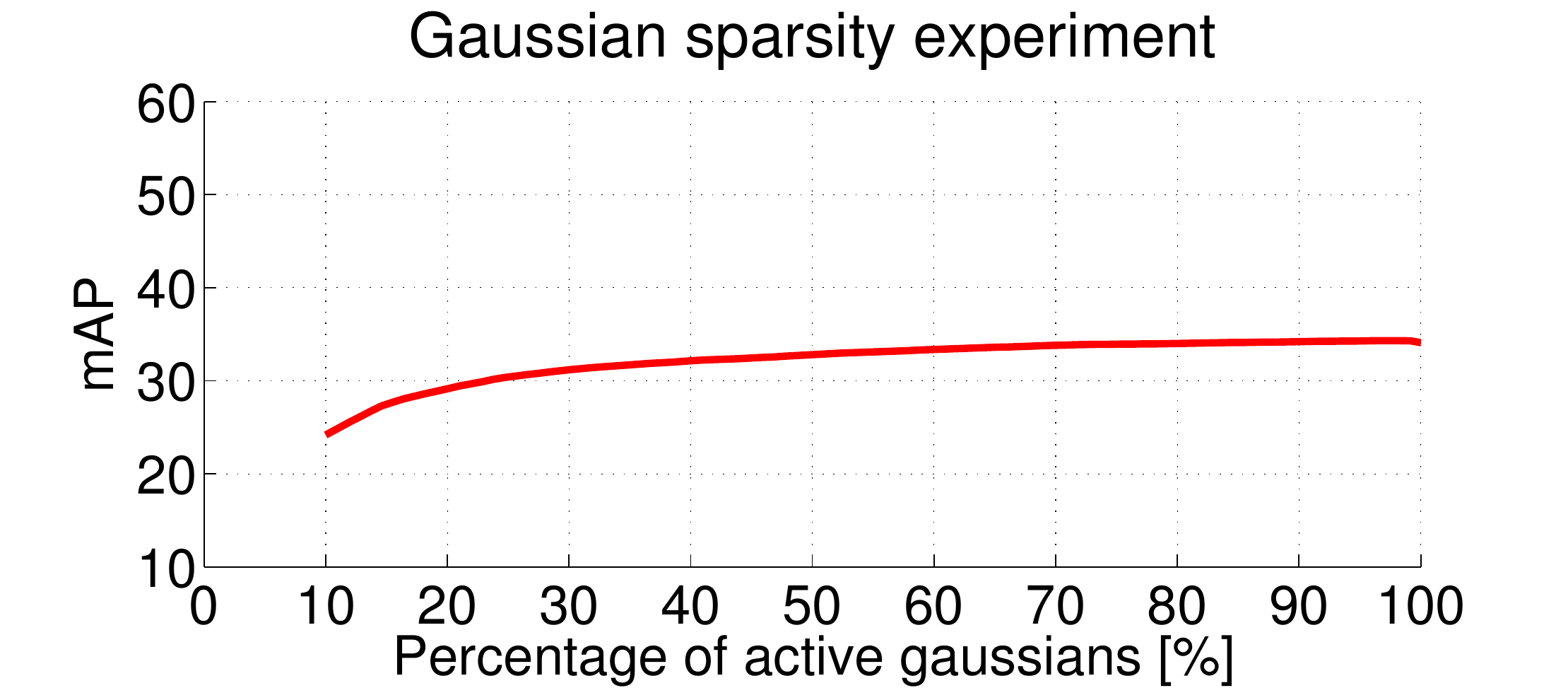}
\end{tabular}
\caption{
Sparsity experiments. Left: as an increasing number of low-scoring patches is removed from the model, detection AP changes very little until 80\% of the patches are discarded (the AP of four representative classes is shown). Right:  detection mAP (across all classes) remains stable by removing up to 50\% of the Gaussian components from the model. See text for details.
}
\label{figure:patchRemoval}
\end{figure}

The experiment was carried on the
development set VOC-07-SMALL. The left panel in Figure \ref{figure:patchRemoval} shows the results.
It is apparent that even after removing about 80\% of the patches with low absolute values of their scores the detector performance remains largely unchanged,
thus confirming our intuition that only a small subset of patches is
actually important for the classifier. We call this property \emph{patch-level sparsity}.

\myparagraph{Sparsity of the Gaussians components}
We have shown earlier that the FV is capable of capturing complex decision boundaries despite making use of a small visual codebook. Here we investigate whether any of these visual words are in fact not informative and can be discarded during detection yielding, among other things, a computational saving. To do so we take advantage of the block structure of the FV.
As explained in Section \ref{sec:fv}, each FV contains $K$ non-overlapping blocks of size $2D$ corresponding to individual Gaussian
components. The idea is then to induce sparsity at the block level when training the model, thereby encouraging the classifier to discard some of these Gaussian statistics.

In order to induce block sparsity, we learn the FV model vector $\bw$ using an SVM but replacing
the standard $\ell_2$ regularizer $\lambda \|\bw\|^2_2$ with the \emph{group lasso} one
$\Omega(\bw)_{\ell_1/\ell_2}$ \cite{yuan2006model}. More precisely $\Omega_{\ell_1/\ell_2}(\bw) = \lambda \sum_{g=1}^K {\|\bw_g\|_2}$, where $\bw_g$ is
the set of weights corresponding to Gaussian $g$ in the FV statistics. Increasing the regularization strength $\lambda$ favors models for which many Gaussians $g$ have $\|\bw_g\|_2=0$, effectively
removing more Gaussians from the representation.
The SVM objective function is optimized using the dual averaging method of \cite{yang2010online}. 
A naive application of this method, however, results in a substantial performance drop as the sparsity of $w$ increases. A possible reason for this loss of performance is that group lasso is capable of selecting useful component, but that the $\ell_1/\ell_2$ regularization is just not competitive with vanilla $\ell_2$. In order to validate this hypothesis, we employ a fine-tuning step in which first group lasso is used to select a subset of useful Gaussians, and then a standard SVM classifier with $\ell_2$-norm regularization is used to learn the final detection model. Perhaps surprisingly, $\ell_2$ retraining recovers most of the lost performance, and is therefore essential to obtain a good sparse model. For example, from the results of experiments on the development set (VOC-07-SMALL) we observed that when using the model which discards 50\% of the Gaussians, there is a decrease of 8.3 mAP (\ie the mean AP computed over all four classes from the development set) if the fine-tuning step is omitted. We think that these observations may transfer to several other applications of group lasso.

Overall Figure~\ref{figure:gaussianSparsity} shows that we can remove
up to 50\% of the Gaussians and still obtain comparable results to the full model. Note that here Gaussians are counted on a per-spatial-bin basis, as they are reused in different ones. As such, the 100\% mark on $x$-axis of Figure~\ref{figure:gaussianSparsity} corresponds to $(4 \times 4+1) \times 64 = 1088$ active Gaussians. Nevertheless, eliminating Gaussian components allows us to avoid accumulating corresponding point-wise FVs during the detection phase yielding a proportional acceleration in detection (note that patches can be quantized once for all detection windows in an image, but accumulation occurs for each candidate window separately).

\section{Summary}
\label{sec:ccl}
\input{conclusion}

\bibliographystyle{plain}  
\bibliography{vedaldi,egbib}
\end{document}

%% file: fisher_vector.tex
\section{Fisher-Vector detection}
\label{sec:fv}

For the purpose of our analysis, we re-implemented the FV object detector described in~\cite{cinbis13segmentation}. 
To focus on the FV representation and avoid confounding factors, 
we did not employ the color descriptors, the contextual rescoring, or the local
feature weighting using masks. 
Instead, we reproduced, and even slightly improved, the ``baseline'' version of their detector. 
We first describe this FV detection pipeline and 
then validate it empirically.

\subsection{Detection pipeline}
The \emph{Fisher Vector} (FV), as used in computer vision applications~\cite{perronnin10improving}, is obtained 
by aggregating the first and second order statistics of local descriptors, here SIFT~\cite{lowe99object}, describing corresponding image patches.
Given a Gaussian mixture model (GMM) with parameters $\mu_k$ (means), $\sigma_k$ (diagonal covariance matrices), and $\pi_k$ (priors),
each $D$-dimensional SIFT descriptor $x_i$ is first assigned to a mixture component $k$ (following~\cite{cinbis13segmentation}
we use hard assignments),
%
%
and then the following first- and second-order statistics are computed:
\begin{equation}
\phi_{i,k}^{(1)} =  \frac{1}{\sqrt{\pi_k}} \frac{x_i-\mu_k}{\sigma_k}, \qquad
\phi_{i,k}^{(2)} = \frac{1}{\sqrt{2 \pi_k}} \left(\left(\frac{x_i-\mu_k}{\sigma_k}\right)^2 - 1\right).
\label{eq:fv}
\end{equation}
Every patch $x_i$ is represented by the concatenation of the component statistics $\Phi_i=(\phi_{i,k}^{(1)},\phi_{i,k}^{(2)}:k=1,\dots,K)$
which we refer to as \emph{point-wise FV}~\cite{chen2013emas}. 
Its dimensionality is equal to $2KD$, 
where $D$ is the dimension of $x_i$ and $K$ is the number of Gaussian components. 
Given an image region (\eg a bounding box) from which we extract $N$ patches,
the final FV $\Phi$ is the average of point-wise FVs: $\Phi = \frac{1}{N}
\sum_{i=1}^N {\Phi_i}$. 


\myparagraph{Weak geometry} 
To incorporate weak geometry in the representation we follow~\cite{cinbis13segmentation} and use a spatial pyramid~\cite{lazebnik06beyond}: each candidate image region is subdivided into  $1\times 1$ and $R\times R$ spatial subdivisions and the corresponding $R^2+1$ FVs are extracted and stacked.


\myparagraph{Normalization}  
As shown in~\cite{perronnin10improving}, the performance of the FV can be
substantially improved by signed-square-rooting (SSR) 
each dimension followed by $\ell_2$ normalization.  The
SSR is used in~\cite{cinbis13segmentation}.  Here we have found that slightly
better performance 
can be obtained by switching to
\emph{intra-normalization}~\cite{arandjelovic13all-about,simonyan13deep}, \ie by applying $\ell_2$ normalization to each individual component statistics after
aggregation:
\begin{equation}
\Phi' =
\frac{1}{\sqrt{K}}
\begin{bmatrix}
\frac{\Phi^1}{\| \Phi^1 \|_2} & ... & \frac{\Phi^K}{\| \Phi^K \|_2}
\end{bmatrix},
\qquad\text{where}\qquad
\Phi^k = 
\begin{bmatrix} 
\sum_{i=1}^N{\phi_{(i,k)}^{(1)}} & \sum_{i=1}^N{\phi_{(i,k)}^{(2)}} 
\end{bmatrix}.
\end{equation}
Note that the $1/\sqrt{K}$ factor guarantees that $\|\Phi'\|_2=1$ when at least
one patch is assigned to each Gaussian. 

\myparagraph{FV detection} 
So far, we have described the FV as a method to compute a descriptor for a
particular image region.
To use this to detect and localize object occurrences, we could apply it in a
sliding window manner over the image. For efficiency reasons, however, we follow~\cite{cinbis13segmentation} and, instead of trying all image subwindows, we limit the search to the ones
enclosing the object region proposals obtained using the algorithm of~\cite{Uijlings13}.

\subsection{Experimental setup}
\label{sec:datasets}


We evaluate our FV pipeline on the PASCAL VOC 2007 
challenge dataset (VOC-07)~\cite{everingham07pascal}, a standard test-bed that
contains 20 categories of animals, vehicles and indoor objects. 
Following~\cite{cinbis13segmentation} we use a small subset of images
from the VOC-07 training set -- referred to as VOC-07-SMALL -- involving only 4 classes to validate the 
parameters of our pipeline. 
This small set is also considered later for the most
computationally-demanding experiments.

We extract $12\times12$ patches with a step size of three pixels at fifteen scales
separated by a factor $1.2$. SIFT~\cite{lowe04distinctive} descriptors are
extracted for each patch. We project and decorrelate these descriptors to $D=64$ dimensions using
PCA. The visual codebook consists of $K=64$ Gaussian components and we use $1\times 1$ and $R\times R$ non-overlapping spatial subdivisions with $R=4$ in the spatial pyramid. This results in the concatenation of 17 intra-normalized FVs, yielding a $17 \times 2KD=139K$ dimensional descriptor, which is then $\ell_2$-normalized again.
We extract around 1500 candidate windows per image using selective
search~\cite{Uijlings13}. 
At train time, we apply 3 rounds of hard-negative mining extracting each time the top two false positive detections for each training image.
At test time, non maximum suppression is applied in order to discard redundant
detections that overlap more than 30\% using the intersection-over-union overlap measure.


 
This pipeline obtains a mean average precision (or mAP) over the 20 classes
of 33.9\% with SSR, which is comparable to the 34.0\% reported
by~\cite{cinbis13segmentation}, and a mAP of 34.6\% with the
\emph{intra-normalization} scheme, thus confirming the positive effects of this normalization.


%% file: conclusion.tex
In this paper, we have shown that the FV detector contains parts in the same manner as a DPM.
In both cases, a fixed number of parts is used to capture the diversity of appearance of an object category. However, this diversity is represented differently: the DPM uses
a mixture of components where each component corresponds to an aspect,
and the FV factors the appearance in a product of multimodal distributions, one for each object part,
encoding implicitly an exponential number of combinations. Both DPM and FV are sparse, although
in somewhat different senses. DPM is sparse at the level of the parts, which are few and max-pooled, while FV exhibits sparsity at the level of part fragments and at the level of the visual vocabulary. For example, while FV has typically more parts than DPM,  80\% of the pooled patches and 50\% of the visual words can be removed
with minimal impact on performance. The latter fact can be used in order to accelerate detection proportionally.
